\documentclass[10pt, sigplan]{acmart}

\usepackage{ifthen}
\usepackage{xspace}
\usepackage{graphicx}

\newcommand{\Title}{Towards a Smart Data Processing and Storage Model}
\newcommand{\SubTitle}{}
\subtitle{\SubTitle}

\newcommand{\Authors}{Ronie Salgado, Marcus Denker, Stéphane Ducasse, Anne Etien, Vincent Aranega\\[2 ex]
RMoD, Inria}

\hypersetup{
	colorlinks=true,
	urlcolor=black,
	linkcolor=black,
	citecolor=black,
	plainpages=false,
	bookmarksopen=true,
	pdfauthor={\Authors},
	pdftitle={\Title}}

\usepackage{xcolor}
\usepackage{textcomp}
\usepackage{listings}
\definecolor{source}{gray}{0.9}
\newcommand{\ct}{\lstinline[backgroundcolor=\color{white},basicstyle=\ttfamily\small]}

\usepackage{listings}
\lstdefinelanguage{smalltalk}{
	basicstyle=\ttfamily\small,
}

\usepackage{xcolor}
\usepackage[normalem]{ulem}

\newboolean{showcomments}
\setboolean{showcomments}{true}
\ifthenelse{\boolean{showcomments}}
	{\newcommand{\nb}[3]{
		{\colorbox{#2}{\bfseries\sffamily\scriptsize\textcolor{white}{#1}}}
		{\textcolor{#2}{\sf\small$\blacktriangleright$\textit{#3}$\blacktriangleleft$}}}
	 }
	{\newcommand{\nb}[3]{}
	 }

\DeclareGraphicsExtensions{.png,.jpg,.pdf,.eps,.gif}

\newcommand{\ie}{\emph{i.e.,}\xspace}

\newcommand{\etal}{\emph{et al.}\xspace}

\newcommand{\seclabel}[1]{\label{sec:#1}}
\newcommand{\figlabel}[1]{\label{fig:#1}}

\newcommand{\lstlabel}[1]{lst:#1}

\newcommand{\lstref}[1]{Listing~\ref{lst:#1}}
\newcommand{\secref}[1]{Section~\ref{sec:#1}}

\DeclareCaptionType{copyrightbox}

\begin{document}

\setlength{\pdfpageheight}{\paperheight}
\setlength{\pdfpagewidth}{\paperwidth}

\copyrightyear{2020}
\acmYear{2020}
\setcopyright{acmcopyright}
\acmConference{IWST '20}{September 29--30, 2020}{Novi Sad, Serbia}\acmPrice{15.00}\acmDOI{10.1145/3139903.3139916}
\acmISBN{978-1-4503-5554-4/17/09}


\title{\Title}

\author{Ronie Salgado}
\affiliation{\institution{Inria, Univ. Lille, CNRS, Centrale Lille}  \city{Lille}
  \country{France}}

\author{Marcus Denker}
\affiliation{\institution{Inria, Univ. Lille, CNRS, Centrale Lille}  \city{Lille}
  \country{France}}

\author{Stéphane Ducasse}
\affiliation{\institution{Inria, Univ. Lille, CNRS, Centrale Lille}  \city{Lille}
  \country{France}}
\email{stephane.ducasse@inria.fr}

\author{Anne Etien}
\affiliation{\institution{Université de Lille, CNRS, Inria, Centrale Lille, UMR 9189 - CRIStAL}\city{Lille}
  \country{France}}

\author{Vincent Aranega}
\affiliation{\institution{Université de Lille, CNRS, Inria, Centrale Lille, UMR 9189 - CRIStAL}  \city{Lille}
  \country{France}}


\begin{abstract}


In several domains it is crucial to store and manipulate data whose origin needs
to be completely traceable to guarantee the consistency, trustworthiness and
reliability on the data itself typically for ethical and legal reasons.
It is also important to guarantee that such properties are also carried further
when such data is composed and processed into new data. In this article we present the main requirements and
theorethical problems that arise by the design of a system supporting data with such
capabilities. We present an architecture for implementing a system as well as a
prototype developed in Pharo.
\end{abstract}





\maketitle

\section{Introduction}
The first objective of this paper is to introduce the need for new object-oriented and
distributed database management with automatic and guaranteed support for the
following three properties or operations on the data system:

\begin{enumerate}
\item Traceability of origins.
\item Automatic verification of integrity.
\item Revocation of data when it becomes invalid, or it becomes private.
\end{enumerate}

We start in~\secref{motivation} by describing three different use case scenarios
motivating the need for this system: keeping track of medical clinical trial data; keeping soft real time
track and monitoring of medical data coming from sensors such as the heart
beat of a person; and the accounting and book-keeping process in a company.
After describing these use case scenarios, we present
a problem statement and a research hypothesis in~\secref{problem-statement}
with the objective of identifying the general principles and requirements that need to be on the foundational object-oriented system to
facilitate the construction of a solution. In~\secref{smart-data-requirements}
we compile a list of fundamental requirements, and in~\secref{fundamental-problems}
we present a list with fundamental problems, risks, along with possible
strategies of mitigation. Finally we present our design and implementation for our smart
data system in~\secref{design-implementation}.

\section{Motivational scenarios}\seclabel{motivation}

\paragraph{Clinical trial data.} Any new medicine or medical procedure needs to
be subject to extensive clinical trials to evaluate its safeness and
efficacy before it can be applied to the general public. This clinical trial
process typically involves several phases, several populations, and several
years of extended controlled experiments. 
All of these controlled experiment processes generate massive
amounts of data whose traceability has to be tightly controlled to
ensure the reliability of gathered data. Tampering with the gathered data should
not be possible. Further posterior analysis, peer reviewal and auditing on the
gathered data may reveal a flaw during certain experiment instances which may
require the revocation of the affected data. This revocation process may
have a cascade effect on complete or partial revocation on the data. In addition
raw data may be used to generate derived data and it is important to know when a derived
data is impacting by the revocation of primary form of data.

\paragraph{Vital sign monitoring system.} The diagnostics and treatment
of several medical disorders such as hyper-tension, hypo-tension, cardiac
arrhythmia, and insulin dependent diabetes may require the constant monitoring
of a patient. This monitoring is carried with an electronic sensor during long
periods of time such as weeks, months and in some cases even
during the whole life of the patient. The ubiquity of some sensors such as a
heart rate sensor present in a watch being weared by an individual makes
constant and daily monitoring feasible for non-diagnostic or treatment purposes.
With the case of hear rate monitoring, a simple scenario involving three
different actors is the following:

\begin{enumerate}
\item The heart rate of a patient is being monitored by a sensor.
\item The family of the patients want to know on whether he or her is alive, dead, or dying.
\item The doctor that treats its patient needs to check a log and a chart of the patient heart rate data once per-week.
\end{enumerate}

In \secref{heart-rate-monitoring} we present a prototype and simulation for this
distributed scenario using our smart data storage and processing system prototype.

\paragraph{Company accounting system.} Many companies need to keep a complete
accounting system on every commercial movement that they perform. The need
for keeping an accounting system comes from two different, but related problem:
keeping track of the everyday operations of the company, and complying with
legally required book-keeping obligations for taxing purposes. In the first
case the company has some flexibility on deciding which additional accounting
records to keep and some specific reports that it might generate
typically for taking commercial and strategic decisions on the company business.
For the case of the taxing purposes, the company needs to keep track of at least
every cash movement flow, by keeping track on each one of its incomes and outcomes,
keep copies of bills, receipts and invoices for each acquisition and sale. Then the company, depending on its jurisdiction, it needs to
aggregate this data to generate periodically different accounting
books in a specific way with a format that is mandated by the law.

\section{Problem statement and hypothesis}\seclabel{problem-statement}

\paragraph{Problem Statement.} From the previous use case scenarios, we derive the following general problem statement:

\emph{How do we model, design and construct a data storage and processing system that
supports automatic traceability of origins, integrity verification, and revocability of invalid
and or private data in a distributed and decentralized environment?}

\paragraph{The Need for Distribution} In the vital sign monitoring scenario the need for distribution and
decentralization is an evident requirement due to the fact that the sensors are
physical distinct machines to the one that is used by the medic and the family
members of the patient. In the other cases, the need for distribution and
decentralization is not that evident, but it stems from the way that large
organizations operate in a distributed geographical scale. For example, large
pharmaceutical companies may conduct their clinical trials in different countries
simultaneously. In the case of the accounting system a large multi-national
may have the need to aggregate and audit the balances of their different local
subsidiaries, in order to keep a global balance for advertising their total
value so that the company can trade its shares in the stock market. For these
reasons, the need for distribution and decentralization can only be ignored in
the case the whole system needs to be executed in a single machine, and there is
also not going to be any need to scale in the future. Otherwise, the system is
gonna require a full redesign and rewriting to add support for distribution
when is needed. On the other hand, a system designed for distribution can always
be executed in a single physical node in case it is required to reduce hardware
cost.

\paragraph{Proposal.} To facilitate the solution of this problem we propose the construction
of a framework in Pharo \cite{Blac09a} around the concept of Smart Data for describing and storing
object oriented data models with the following capabilities:

\begin{enumerate}
\item Automatic support for traceability of origins.
\item Guaranteed structural data consistency via immutability enforced in transaction
boundaries for specific historical data snapshot.
\item Automatic data replication and distribution is always possible, and easy for scenarios with a
fixed and known data distribution topology.
\end{enumerate}

In the next section we discuss the actual requirements for our smart data
processing and storage system by defining these requirements.

\section{Smart data storage system requirements}\seclabel{smart-data-requirements} 

\subsection{Traceability of origins}

\paragraph{Definition of Traceability.} We define traceability of origins as the
ability of knowing where each piece of data in a data base comes from. A more
concrete definition means that for each datum $D$ that is present in a database in a
specific period of time we should be able to answer at least the following
questions:

\begin{itemize}
\item \emph{Who wrote} the datum $D$?
\item \emph{When} the datum $D$ was \emph{written}?
\item With \emph{which and whose} permissions was the datum $D$ \emph{written}?
\end{itemize}

\paragraph{Origin as Metadata.} The first important observation of these questions is that they are all about
requesting a specific \emph{metadata} about the datum $D$. The presence of this
metadata in this context is orthogonal to the actual data, so a desirable
property is to keep the specification of this metadata completely separate from
the specification of the actual domain specific data model.

\paragraph{Transactions.} The second important observation is that
all of these questions has the verb \emph{write}. For our purposes, we are
going to define a \emph{transaction} as a \emph{contained} and \emph{isolated}
process where a sequence of \emph{reads} and \emph{writes} is performed on the
data stored in a database. To simplify our design we are also going to mandate
that data accesses happen during the context of a \emph{transaction}. A
transaction might end on a \emph{commit} where the data is actually written
into the data store, or it might end in a \emph{rollback} where the new data
is completely discarded. During a transaction \emph{commit}, the newly written
data has to be serialized and transmitted into the final persistent data store,
which may be a single disk drive, or even a replicated cluster. The important
aspect here is that transactions introduce synchronization points where a
specific version of the data is \emph{read} and \emph{written}.

\paragraph{Transaction as the Origin.} Since any piece of data can only be
\emph{written} during the context of a transaction, then it becomes evident that
all of the data written in the system has a single transaction as its source
of origin. This implies that the problem of traceability of origin can be
solved by just tagging transaction themselves with additional metadata that is
part of the transaction life cycle. In other words, the previous questions
about the data origin can be re-phrased in the following ways:

\begin{itemize}
\item \emph{Who} initiated the transaction $T$ that wrote the datum $D$?
\item \emph{When} was the transaction $T$ that wrote the datum $D$ committed?
\item With \emph{which and whose} permissions were used to initiate and commit the transaction $T$ wrote the datum $D$?
\end{itemize}

\paragraph{Transaction Context Specific Data.} Since adding this metadata to the transaction itself is completely orthogonal
to the actual data domain, then we propose that the problem of traceability of
origin can be solved by using transaction context specific data. We propose this
mechanism as an elegant and more robust alternative than to actually model the
concept of origins in the data model itself.

\subsection{Integrity verification}
The capability of performing \emph{integrity verification} implies that we have a predicate function $I$ that
receives at least a subset of the whole data set, and answers true or false on
whether the subset of the data set complies with some requirements on data integrity. For our
purposes, we decided to define our predicate $I$ as the simultaneous compliance of the
following three related predicates: structural consistency on the data ($S$);
domain defined constraints ($D_c$), and bit-level data correctness ($B_c$). In
algebraic form, this is summarized in the following way:

\[
I(D) = S(D) \land D_c(D) \land B_c(D)
\]

We also define each of these different components of the integrity verification
predicate as the simultaneous compliance of multiple boolean predicates. We
define these three predicates as grouping of different aspects on the concept
of data integrity.

\paragraph{Structural Consistency.} We define \emph{structural consistency} as the
\emph{correctness on the organization} of the stored data. The main factor
required for enforcing this is \emph{type checking correctness}. For example,
this implies that it should be impossible to write a \emph{date} in a place
where a \emph{home address} is required. In addition to type checking, we
also define the requirements on the data topology as being part of its
structural consistency requirement. For example, it should not be possible
to have a cyclic graph of objects in a location where a tree is required. Simple
cases of these requirement on the topology might be encoded as type checking
requirements, but non-trivial cases might need additional predicates and
graph traversal algorithm for testing.

\paragraph{Domain Defined Constraints.} In the $D_c$ predicate we group all of
the boolean predicates that specify additional domain specific
constraints that are mandated by the application. These predicates are
completely user defined and they encode the invariants and the contracts that
their data model always has to comply. For example, in double-entry
accountability each book keeping entry that represents an income from one
account must have a corresponding outcome entry from another account.

\paragraph{Bit-level Correctness.} Data has to eventually be serialized and
transmitted via some link to its final persistence storage medium. This
transmission and storage process may be imperfect and suffer of data losses. We
define as \emph{bit-level correctness} to the data integrity aspects that arise during
this serialization, encoding and storage process. The standard mechanism to
detect and repair this problem consists on the usage of checksum, and error
detection and corrections codes. Stronger mathematical guarantees on this
aspect can be provided through the usage of cryptographic hashes. The
introduction of cryptography on this level also provides the opportunity of
using a \emph{digital signatures algorithms} (DSA) or
a \emph{message authentication code} (MAC) to also guarantee the
authenticity of the data. The capability of adding cryptographic validation at
this level also allows one to add an additional layer to detect and prevent data
tampering by unauthorized parties.

\subsection{Revocation}

We define data revocation as the process of turning previously valid data
into invalid data, and automatically reacting on the dependent data by
repairing it or also marking it as invalid data. We identify two main reasons
for revocating data: data that was previously valid is not valid anymore
because new sources of information arrived and contradicted the
previous validation of data; and the need for destructing or hiding
information that is not needed anymore and has become private.

\paragraph{Revocation as Data Invalidation.} After data is committed into a data
store, new information may be added that may invalidate at least partially
the modified data during a previous transaction. For example, in the case of
the clinical trial, it may be determined after an auditing process that some
instances of a controlled experiment for a drug are invalid due to a mistake
committed during the execution of those specific experiment. Once the flawed
experiments are detected, it becomes necessary to re-generate the reports that
depend on this experimental data by excluding the data from those experiments,
and in some extreme cases it may even invalidate an actual approval for a
drug. As another example, in the case of a company accounting system, sometimes
sold products are refunded, so that specific sale becomes invalid.

\paragraph{Revocation as Data Destruction.} Data may need to be destructed or its
access may be needed to make private due to many possible reasons. One important
motivation for destroying data is the need to preserve the privacy of individuals
and respect laws on data protection and privacy such as the GDPR. These laws
also allow the individual the right for having their private data removed from
electronic services. This means that they may be even necessary by a mandate by law to
completely destroy the private data from the storage system. In some cases it might be
enough to replace the private data by completely anonimized data, however
proper automatic anonimization of data is a difficult problem because it is also
necessary to provide the guarantee of making it impossible to de-anonimize the
data by correlating multiple data sources.

\paragraph{Cascade Effect.} Revocated data may exist in a dependency to additional
data which may also need to be revocated. In the case of revocated data due
to invalidation, it might be enough to mark the data that depends on it as invalidated
and generate a new version of it to re-validate the data. This may
produce a cascade sequence effect of invalidation and re-validation. If the set
of affected data forms a directed acyclic graph (DAG), then the problem of re-validating this data is a
just a matter of performing a topological sort on the directed graph, and then
execute the single sequence of invalidation and revalidations. If the affected
data instead has cycles, it may not even be possible to completely validate
it, and this process may even have a catastrophic effect. For example, a
directed cyclic graph with two vertices $V = \{a, b\}$, and two edges $E = \{(a,b), (b, a)\}$.
In this graph the invalidation and re-validation of $a$ may trigger the
invalidation of $B$, whose revalidation may trigger a new invalidation of
$A$. In the general case, the invalidation and revalidation process may  never
reach a steady state where it completes. This means that in the cyclic case it
is required to find a strategy that either:

\begin{itemize}
\item Ensures that a revalidation steady state is always reached.
\item Detect this oscillatory process, forces them into a new state, or destroys them.
\item Perform re-validation in a lazy process. This mitigates the nasty meta
    stability issues by bounding computations during revalidation into single
    slice of a bread-first search style graph traversal.
\end{itemize}

As for the case of revocated objects due to destruction, if the cascade effect
of revocation involves the destruction of objects that depends on it, it
becomes evident that the consequences might be completely catastrophic. If on the
other hand it only involves a process of invalidating and re-validating the
affected data.

In our current prototype we have not yet implemented any kind of support for the
data revocation capability yet, and we still have to decide on how to properly model
and express these two different cases of data revocation.

\subsection{Concurrency, Distribution and Decentralization}

Concurrency and distribution are nowadays two central requirements on any
real world data storage and processing system that needs scale to large
numbers of connections and simultaneous transactions. This implies that our
smart data storage and processing system has to be designed at the architectonic
level to support concurrent transactions, and to at least facilitate the
possibility of replicating and propagating data.

Another desirable property is the ability to decentralize the management structures and
authorities that are involved in the operation of a distributed system. This
desire to support decentralization comes from two sides: real work requirements
on how data is produced, stored and shared by different real world organizations;
the need of having robustness against failures, and even against some malicious
attacks on a data infrastructure by one owner.

\section{Fundamental Problems, Risks and Mitigations}\seclabel{fundamental-problems}

\subsection{Concurrency and Distribution Issues}
Since the support for concurrency and distribution is an important and
desirable property, this means that our system has to be designed to avoid, and
if possible, to make impossible the main threats of concurrent and distributed
system that might bring the whole system down. These crucial well known threats
that need to be at least mitigated are the followings:

\begin{itemize}
\item Deadlocks: mutual locks that are never released.
\item Livelocks: mutual actors that are waiting for each other.
\item Race conditions: conflicting read-modify-writes on shared state.
\end{itemize}

\paragraph{Deadlock.} Deadlocks are caused by two different processes that are taking two different
locks in a different order. Deadlocks are solved by holding these locks always
on the same order. The presence of explicit transactions helps on mitigating this
issue since transactions induce an ordering constraints on the operations, which
also provides an opportunity for taking the locks in the proper order.

\paragraph{Livelock.} Livelocks typically happens on the presence of two synchronous processes that are
waiting for each other, so they never end waiting and doing an actual useful
work. One solution to this problem is on having actors that only communicate via
asynchronous communication. This also introduces the problem that some concurrent
and distributed processes need to be designed and implemented in terms of state
machines. In our smart data storage and processing system we are still not
modeling and supporting properly actors, so for now we are not attacking this
problem yet.

\paragraph{Race Condition.} Race conditions are the product of conflicting
read-modify-write (RMW) operations on a shared state by two or more simultaneous
processes. The traditional way for preventing race conditions consists on
introducing locks. Another way of preventing race conditions consists on not
sharing data that is simultaneously being written at the same time at all. One
way of enforcing this property is by making all data immutable, which is the
main property that is guaranteed by pure functional programming languages. The
problem of this approach is that it prevents conducting traditional object-oriented programming which is plagued of mutation everywhere. In the next
subsection we discuss a hybrid approach where we enforce immutability on certain
points in our data model to simplify our data storage model, and to also
retain the convenience of mutable object state oriented programming for data
manipulation.

\subsection{Mixing Mutability and Immutability}
\paragraph{An Apparent Contradiction.} Since several concurrency issues are solved
by having immutable data, then it becomes desirable to enforce it as a property
on the stored, transmitted and replicated data. However, since we are
implementing and using Pharo, a purely object-oriented programming language
that is not designed to deal and ensure immutability at every moment unlike a
purely functional language such as Erlang \cite{hebert2013learn}, then the need
to enforce immutability at the language level looks like an unnecessarily restrictive constraint.

\paragraph{Serialization as a Conversion.} Fortunately, data is always manipulated in the context
of a transaction. This means that we only ``perform mutation'' in a transient
object-oriented representation of the data that is serialized for persistence
at the end of the transaction. For final transmission
or storage, the data has to be converted into a linear string of bits through a
process of serialization. If we decide to always serialize the complete modified
objects, then this serialized version of the object is by definition
immutable. However, we lose this immutability property on the serialized data
if we simply replace the old serialized version of the object with the new
version. If instead of replacing the old version, we decide to keep both versions,
and to always append data at the end of data store, then we have an immutable
data persistence system. This means that there are no data races at the
object persistence level because we are only storing immutable data by appending
it to existing one.

\paragraph{Transaction Delimited Immutability.} In other words, we use read
operations during the context of a transaction for deserializing a snapshot of
the database and converting an immutable version of an object into a mutable
object that is local to the transaction context. At the end of a committed
transaction, we serialize the mutable objects into an immutable representation,
and we submit this representation into the data store. From the point of view of
the final data store, there are only immutable operations on it. But the user
sees an object-oriented interface that looks like holding convenient mutable
objects.

\subsection{The CAP theorem}
The Brewer's CAP theorem \cite{simon2000brewer, gilbert2002brewer} is a well
known result and fundamental problem in distributed computing. This theorem
states that it is impossible to construct a distributed system that guarantees
all of the time the following three properties:

\begin{itemize}
\item Consistency (C).
\item Availability (A).
\item Tolerance to network partitions (P).
\end{itemize}

The CAP theorem also states that it is possible to construct distributed system
that guarantees at most two of these properties at the same time. Under this
view, it is typically said that traditional relational database management
systems (RDBMS) are designed to guarantee consistency and availability through
ACID transactions. 

In the case of distributed systems, all of them have to actually support the
P component of theorem \cite{hebert2013learn}. This means that a distributed
system might only be able to choose providing an additional guarantee for
either Consistency or Availability \cite{hebert2013learn}. Also, for different
components of a distributed system it might be desirable to make different
choices between keeping A or C \cite{hebert2013learn}. For example, the
accounting department of a seismographic research institute needs to guarantee
consistency, but its seismic data reception, storage and aggregation has to
instead guarantee the availability.

\subsection{ACID Transaction}
ACID is the acronym with the main properties that have to be fullfilled by
transactions that are used in traditional relational databases. These properties
are the followings:

\begin{itemize}
\item Atomicity: the whole transaction cannot be divided.
\item Consistency: the database only ends on a valid state at the end.
\item Isolation: transactions runs as if there was only a single transaction running at a time.
\item Durability: once the transaction is finished and committed, its result is stored in a permanent storage medium.
\end{itemize}

In terms of the CAP theorem, ACID transactions are used for guaranteeing the
Consistency attribute in the database. For this reason, it is mandatory for our
system to be able to support this kind of transaction. By the previous definition
of transaction, the ACID properties in our system can only be violated by running multiple
concurrent transactions that are either: conflicting by themselves, or the
summation of both transactions invalidates an integrity constraints on an
object depends on the modified objects by the transaction. With our system we
can support ACID semantics by having a global synchronization mechanism that
prevents conflicting transactions, and by also triggering the automatic
repairing of invalidated data as a consequence of these transactions.

\subsection{BASE Transaction}
BASE is an alternative to the traditional ACID semantics that  instead of preserving the C property of the CAP theorems preserves the Availability \cite{pritchett2008base}.
The objective of BASE transactions  is to preserve the Availability guarantee by
sacrificing the Consistency property from the CAP triangle. The definition of
the BASE acronym is the following:

\begin{itemize}
\item Basically Available: the A part of the CAP theorem.
\item Soft state: lack of consistency guarantees.
\item Eventual consistency.
\end{itemize}

Eventual consistency means that after some time, the data on these systems
converges into a stable value, and reads will always produce the same
value \cite{vogels2009eventually}. One mechanism to guarantee eventual
consistency is by merging the updates performed by two conflicting
transactions \cite{shapiro2011conflict}\cite{hebert2013learn}. However, there
are some requirements on how this merge operation has to be defined to
achieve eventual consistency \cite{shapiro2011conflict}. It has been shown that
simple and trivial case that always achieves eventual consistency is when the
merge operation is commutative \cite{shapiro2011conflict}.

In our current infrastructure, we still have not decided on a proper way to
model and provide support for these high-availability distribution scenarios where
BASE semantics are required. However, the ability of being able to structure data in
terms of historical versioned immutable structured provides a simple degradated
conflict resolution mechanism. This mechanism is to represent the latest version
of the conflicting objects as a \emph{set} that contains the versions that have
a conflict. With this strategy no data is ever lost, but the conflict has to be
solved explicitly by a new transaction, which could in the worst case be
manually initiated merge procedure by a human individual. This conflict
resolution mechanism is exactly the same mechanism that is used by software
versioning control systems such as Git to solve conflicts.

\subsection{Consensus reaching algorithm}
Reaching consensus is a fundamental problem on any distributed system. The
ability to reach consensus is crucial for guaranteeing correctness on the data
that is computed by a distributed. Unfortunately, there are several known
instances where it has been proven the impossibility of reaching
consensus \cite{fischer1986easy}. This implies that the design decisions that
we make for constructing our smart data storage and processing system has to be
aware of the existence of these impossibilities. This also implies that measures
have to be taken for detecting and mitigating these situations where a
consensus cannot be reached. In the cases where consensus can be reached, the
implementation should provide mechanism to reach it by using existent algorithm
\cite{cortes2008distributed}. This problem can only be ignored in the cases
where there is zero data distribution replication that is meant to improve
fault tolerance.

\section{Smart Data System Design}\seclabel{design-implementation}
\paragraph{Layered Architecture.} To facilitate the design, and to
enforce separation of concerns of the different aspects of our problem, we decided
to construct our smart data storage and processing system in terms of three
layered processes: physical data representation, transaction context processing,
and domain specific modeling and processes.


\subsection{Physical Data Representation}

\paragraph{Key-Value store for Physical Persistence.} In the lowest level we
eventually need to persist the serialized bits of objects in a persistent data
store. As a simple and standard solution to this problem we decided to use a
key-value store \cite{han2011survey} for our lowest level persistence layer. In abstract terms, it
means that we have a single large dictionary whose key is an ID for retrieving
a specific instance of a serialized object. A key-value store offers the
advantage of being easy to implement in a transient and in-image memory
only by using a single \ct{Dictionary} instance, and a single $Semaphore$ for
mutual exclusion and ensuring atomic read and writes of serialized object
data. Having this abstraction also allows one to use different backends, such as:

\begin{itemize}
\item Use an existent RDBMS such as MySQL \cite{mysql2001mysql} only for final disk persistence. A
relational database with a single table of two columns is enough for this
purpose. This takes advantage of the existent database optimizations for storing
data on disk in an efficient and safe way.
\item Use another document based, NoSQL \cite{han2011survey} database such as
MongoDB \cite{mongodb2014mongodb}. These NoSQL databases are typically designed
around this very same abstraction of a key-value data store \cite{han2011survey}.
Since many of these existent databases already have support for data
replication \cite{han2011survey}, their usage might also help on this aspect,
but we still need to implement an additional layer of synchronization if we
want to enforce ACID constraints in some transactions.

\end{itemize}

\paragraph{Document Based Data Encoding.}
A good definition for a document database is the following one given by Han \etal:

``Document database and Key-value is very similar in structure, but the Value of document database is
semantic, and is stored in JSON or XML format.'' \cite{xie2014salt}

This means that databases of this kind such as MongoDB \cite{mongodb2014mongodb}
can be used for storing data structured in terms of a document object model (DOM). The JSON
structure for text based object serialization tends to be preferred over XML for
reasons of simplicity, and better performance due to shorter documents
\cite{maeda2012performance}. JSON is a text format, but there are several
binary formats that are compatible with JSON, faster to parse, and directly
supported by databases such as BSON \cite{specbson} and MessagePack
\cite{furuhashi2013messagepack}. The existence of these compact and binary
alternatives that follow the exact same model of JSON means that using this
same model for structuring and serializing objects is a safe design decision.

A JSON document is composed of a single dictionary object. An object in JSON
can have one of the following recursively defined structure:
\begin{itemize}
\item An atomic literal value:
    \subitem Numbers
    \subitem Strings
    \subitem Boolean values
    \subitem Nil.
\item An array of objects.
\item A dictionary of objects. Keys are always strings, but values can
    be any kind of object, including other dictionaries and arrays.
\end{itemize}

The subset of JSON objects composed by atomic literal values and arrays can be
concisely expressed in Pharo syntax in terms literal arrays through
the $\#(1 2 (true false)$ syntax, and in terms of array construction with
the $\{1. 2. \{true . false\}\}$ syntax. Unfortunately, in Pharo we do not
have a special syntax for defining dictionaries. However, a simple and concise
way to specify a dictionary in Pharo is to use successive pairs of keys and
values. These arrays of key-value pairs can be passed to the \emph{Dictionary newFromPairs:}
method for constructing objects. For this reason, and to facilitate writing
unit tests of our object serialization process, we decided to implement our
serialization process in terms of these arrays, but in our deserialization
process we are in fact converting them into dictionaries to relax the
requirement of having the same order on the encoded fields all of the time.

Another advantage of encoding the serialized data in terms of these document
formats, is that they also allow generating a canonical encoded bitstream
of the serialized data. This canonical encoding is a deterministic definition
on how to encode the data to obtain always the same result. Having a canonical
encoding form is crucial for applying cryptographical algorithms for guaranteeing
bit-level data integrity and authentication.

\paragraph{The Problem of Nominative IDs.} One important observation is that
is not enough to just store directly serialized objects into the data store. If
we store a serialized graph of inter-dependent objects directly by using the
nominative ID (\ie the name) of each object in the graph as the keys in the
key-value store, then it becomes impossible to modify a single object without
having to simultaneously re-validate and modify all of the objects that depend
on it. In other words, by associating directly the name of the objects with its
content we lost any possibility of having an immutable and an append only store
of serialized objects. We solve this problem by using versioned object ID. We
define a versioned ID as an ordered pair composed of the nominative object ID,
and a version ID number. During serialization of object graphs, we replace object
references by a reference into a specific version of the object. However, it its
still desirable being able to obtain the latest version of an object from just
its ID. It is also a desirable property to be able to navigate the whole
history of a stored object. We achieve such properties by keeping a single separate
\emph{mutable} mapping from an object nominative ID into its last versioned
ID. In addition each serialized versioned object has a reference
to its previous version. This means that different versions of an object
maintain a linked-list data structure that allows support for arbitrary
navigation on the history of an object.

\paragraph{Atomic Values, Entities and Roles.} We define three different kinds
of objects that need to be manipulated and persisted through our smart data
storage and processing system: \emph{atomic values}, \emph{entities} and \emph{roles}.

\begin{description}

\item \emph{Atomic values.} We define as
an atomic value any objects whose individual internal components are not
separately versioned and traced for origins. These are the indivisible units of
data such as a number, a string, a boolean, a date, or even some aggregate
objects such as street address. Versioning and origin traceability of atomic
values is conducted on the whole value itself. The only requirements on atomic
values is being able to serialize-deserialize them, and to evaluate their
integrity constraints.

\item \emph{Entities.}
Entities are objects which have complete historical versioning and
traceability of origins. Unlike atomic values, the individual components of an
entity are also versioned and traced. For this reason, we define the internal
persisted state of an entity as a composition of different atomic values. For
reasons of convenience, we define the different components of an entity by
using slots \cite{verwaest2011flexible}. The usage of these slots has the
following purposes:

\begin{itemize}
\item Replace the persisted variables by value holders that intercept reads
    and writes. These are used for tracing modified objects, and automatically
    tagging the modified data with its new origin. By replacing the slots with
    a value holder, each read and write into it becomes a message send. The
    origin of the values only change when the write message is received. On the
    processing of a slot write message, the data originator in the value holder
    is compared with the data originator of the current transaction context. If
    the data originators do not match, then the originator of the value holder
    is changed into the new one, and the object version id is updated to be
    the current in memory version. The interception of reads is used to implement
    lazy deserialization of referenced objects, which allows to avoid reading
    and deserializing the whole object graph from the data store.

\item Make explicit the database schema, and the types of the stored object. Typing
    allows enforcing the type checking integrity constraints.
\item Provide automatic serialization and deserialization of objects. Each slot
    knows its type, and each type knows how to serialize and deserialize its data.
\end{itemize}

\item \emph{Roles.}
We define as a role as any object that can act as a \emph{data originator}\footnote{We
might have just call them \emph{data originator}, but we think that the term
\emph{role} is more adequate since it can be seen as an analog to a real world
individual that is full-filing a specific social role. In a future version we may replace the \emph{concrete role} term with the \emph{data originator} term.}. Many domain
specific roles are also entities, and we call them \emph{role-entities}. But they are
some special roles such as the super-user administrator that is required for
creating the initial roles in a data store that we are not modeling as entities.
For this reason, we construct an object that we call \emph{concrete role} that
represents the act of exercising a role. We mandate the presence of a concrete
role for being able to actually initiate a transaction, and we use these
concrete roles for actually tagging the origin of a transaction.
\end{description}


\subsection{Transaction Processing}

\paragraph{Transaction Life-Cycle.} The life-cycle of a transaction is composed
of three parts: the beginning of the transaction, a sequence of read-modify-write
operations in the transaction context, and the end of a \emph{commit} or
an \emph{abortion} that originates a rollback. \lstref{transaction-example} shows an example
of how a transaction looks like in our system.

\begin{figure}[!htb]
\centering
\lstset{language=smalltalk,caption={Example code of a transaction}, label=\lstlabel{transaction-example}}
\begin{lstlisting}[frame=single]
dataStore withRole: hospitalService doTransactionWith: [ :transaction |
	patient := SMDPatient createWithID: 'Patient1'.
	patient names: 'John';
		surnames: 'Doe';
		birthDate: (Date year: 2000 month: 1 day: 1);
		address: '25 av marechal foch'.
	transaction commit.
].
\end{lstlisting}
\end{figure}

We represent the sequence of read-modify-write operations in as
just a standard Pharo block. We also mandate the requirement of having a specific
concrete role for starting a transaction. This concrete role represents the
origin of the transaction which allows enforcing traceability of origins.

The \ct{withRole:doTransactionWith:} method takes care of all of the
transaction management required book-keeping. The most important action of this
method is to store the active transaction context in a process local variable.
This active transaction context is used by the automatic data originator tagging
machinery through the interception of value writes as described in the previous section.
This interception of writes is also used for compiling a list of objects that
need to be serialized and written to the data store at the end of the
transaction (\ie constructing the transaction log) Another important usage of
this transaction context is on the implementation of automatic object reads and
deserialization on-demand through the interception of reads.

We implement the explicit transaction \ct{commit} and \ct{abort} operations
as the signaling of specific exceptions. The exceptions that are raised during a
transaction block are all caught. If the caught exception is an explicit
\ct{commit} signal, then the modified objects during the transaction are
serialized and submitted for persistence into the data store. All of the other
cases generate a transaction rollback operation, which in the current
implementation version is implemented by simply ignoring the modified objects.

\subsection{Domain Specific Modeling}

Once having these basic building blocks, the next step is to actually model
domain specific data and processes that are actually required by a specific
application.
These objects can be modeled by creating additional definitions of
entities, roles, and in some cases, even new atomic value types.
In our implementation, such definitions can be created by just defining new subclasses.
We present a concrete example of this in~\secref{heart-rate-monitoring}.

\section{Heart Rate Monitoring Demo}\seclabel{heart-rate-monitoring}

With our smart data storage and processing system we constructed for testing it
a simplistic scenario. This scenario is composed of three different actors:

\begin{enumerate}
\item A heart rate sensor that monitors a patient in real time.
\item The family member of the patient that want to know on whether he is still alive.
\item A medic that needs to check once per week the heart rate of the patient.
\end{enumerate}

\paragraph{Non-distributed Implementation.} For reasons of simplicity, we implemented this
scenario in terms of three concurrent processes running inside a single image.
In other words, our current demo is not distributed, but it still has all of the
transaction, object serialization, and traceability of origin, and automatic
object history versioning support capabilities. For this demo we are using a
single transient in-memory data store that is implemented through the
combination of some Pharo dictionaries and semaphores for mutual exclusion.

\paragraph{Data model.} In this scenario, we have the following four different
role-entities:

\begin{enumerate}
\item The patient who is also an individual. See \lstref{patient-entity-definition}.
\item The medic that treats the patient. The medic is an individual. See \lstref{medic-entity-definition}.
\item An individual who is a family member of the patient. See \lstref{individual-entity-definition}.
\item The heart rate sensor itself. See \lstref{heart-sensor-entity-definition}.
\end{enumerate}

\begin{figure}[!htb]
\centering
\lstset{language=smalltalk,caption={Definition for the \emph{patient} entity}, label=\lstlabel{patient-entity-definition}}
\begin{lstlisting}[frame=single]
SMDIndividual subclass: #SMDPatient
	slots: {
		#medics => SMDMedic set .
		#heartRateSamples => SMDHeartRateSample set }
\end{lstlisting}
\end{figure}

\begin{figure}[!htb]
\centering
\lstset{language=smalltalk,caption={Definition for the \emph{medic} entity}, label=\lstlabel{medic-entity-definition}}
\begin{lstlisting}[frame=single]
SMDIndividual subclass: #SMDMedic
	slots: { #patients => SMDPatient set }
\end{lstlisting}
\end{figure}

\begin{figure}[!htb]
\centering
\lstset{language=smalltalk,caption={Definition for the \emph{individual} entity}, label=\lstlabel{individual-entity-definition}}
\begin{lstlisting}[frame=single]
SMDRoleEntity subclass: #SMDIndividual
	slots: {
		#names => SMDStringType .
		#surnames => SMDStringType }
\end{lstlisting}
\end{figure}

\begin{figure}[!htb]
\centering
\lstset{language=smalltalk,caption={Definition for the \emph{heart rate sensor} entity}, label=\lstlabel{heart-sensor-entity-definition}}
\begin{lstlisting}[frame=single]
SMDRoleEntity subclass: #SMDHeartRateSensor
	slots: { #patient => SMDPatient }
\end{lstlisting}
\end{figure}

The main objective in this scenario is to collect a set of heart-rate samples.
This means that we also need to define how a heart-rate sample looks like. Since
each one of these samples is an atomic object by itself, then it makes sense to
model it as an aggregate object with two fields: the timestamp of when the
sample is measured, and the number of beats per minute measured at that time.
 \lstref{heart-rate-sample-definition} shows the use of the
class \ct{SMDCompositeValue} to define this composite value object.

\begin{figure}[!htb]
\centering
\lstset{language=smalltalk,caption={Definition for a heart rate sample}, label=\lstlabel{heart-rate-sample-definition}}
\begin{lstlisting}[frame=single]
SMDCompositeValue subclass: #SMDHeartRateSample
	slots: { #timestamp => SMDDateAndTimeType .
		     #beatsPerMinute => SMDFloatType }
\end{lstlisting}
\end{figure}

\paragraph{Data Store Creation and Initialization.} At the beginning of this demo, it is
required to define the set up of the database by creating actual
instances of these roles, and connect them in the context of a transaction. For
this demo, the code required for creating the data store, and performing this
initialization is given in \lstref{heart-rate-initial-population}.

\begin{figure}[!htb]
\centering
\lstset{language=smalltalk,caption={Data store creation and population}, label=\lstlabel{heart-rate-initial-population}}
\begin{lstlisting}[frame=single]
store := SMDTransientDataStore new.

"Create the individuals"
patient := store
	withRole: SMDSuperUserAdminRole
	getOrCreateRole: SMDPatient withID: #Patient.
son := store
	withRole: SMDSuperUserAdminRole
	getOrCreateRole: SMDIndividual withID: #PatientSon.
medic := store
	withRole: SMDSuperUserAdminRole
	getOrCreateRole: SMDMedic withID: #Cardiologist.

"Create the sensor role"
store withRole: medic doTransactionWith: [ :trans |
	heartBeatSensorRole := SMDHeartRateSensor
			getOrCreateWithID: {patient fullId . #watch }.
	heartBeatSensorRole patient: patient.
	patient medics add: medic.
	medic patients add: patient.
	trans commit.
].

\end{lstlisting}
\end{figure}

\paragraph{Spawning the Actors.} Once the data store is initialized,
the next step consists on actually spawning the different actors that need to interact as shown in~\lstref{actor-spawning-process}.

\begin{figure}[!htb]
\centering
\lstset{language=smalltalk,caption={``Actor'' spawning code}, label=\lstlabel{actor-spawning-process}}
\begin{lstlisting}[frame=single]
"Create the sensor"
sensor := SMDHeartRateSensorProcess new
	dataStore: dataStore;
	role: heartBeatSensorRole;
	yourself.
sensor start; openUI.

"Create the status UI"
SMDHeartRateStatusUI new
	dataStore: dataStore; role: son; patient: patient;
	openInWindow.

"Create the medic UI"
(SMDHeartRateMedicUI on: {dataStore . medic}) openWithSpec.
\end{lstlisting}
\end{figure}

Unfortunately, currently we have not yet defined a proper
framework for actors that have to interact with one of our data stores. For this
reason, we are currently using the term actor as a completely ad-hoc definition
for any distributed process that performs transactions. We are calling them
actors in analogy to the actors of a theater play. In \secref{limitations} we
discuss the limitations of this ad-hoc approach.

\begin{figure}[htb]
\centering
\includegraphics[width=0.5\textwidth]{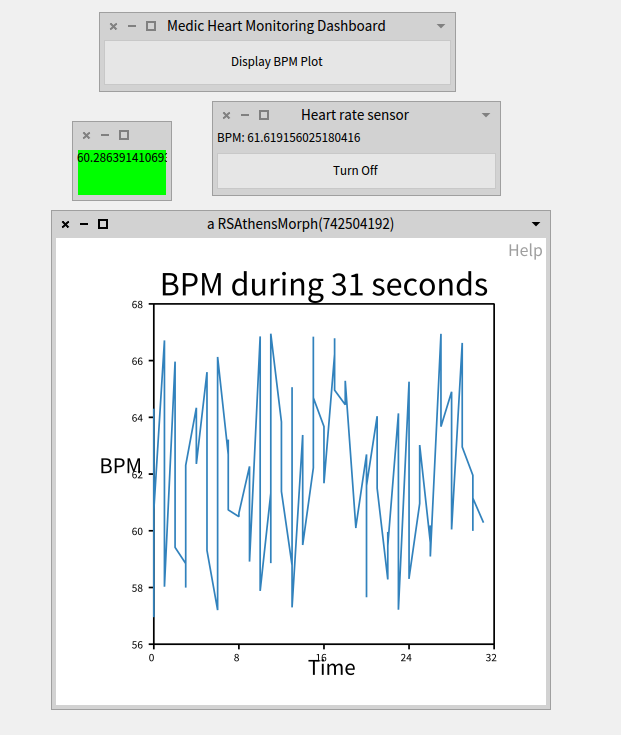}
\caption{Screenshot of the heart rate monitoring demo.}
\figlabel{demo-screenshot}
\end{figure}

\paragraph{Sensor Implementation.} The heart rate monitoring sensor actor is the
physical sensor that takes care of actually sampling the patient heart beat. A
simplistic implementation for this sensing process could be to read a sample, and
then immediately perform a transaction for transmitting the samples into the
data store. The main problem of this simplistic approach is performance because
of two important sources of overhead: the overhead of setting up a transaction,
and the overhead of establishing a network connection in the case of network
distribution. In addition to this problem, samples may need to pass through
electronic and digital filters to actually recover the heart beat
sensing data. For this reason, it is desired to accumulate batches of samples
during several seconds, or even minutes and then perform a transaction to
submit the complete batch of samples. The objective of this batching process is
to reduce the frequency of transmission from the order of milliseconds which
might be even less than the actual latency of the network, to an order of
magnitude that is greater to the normal latency of the network. One important
objective of this is to avoid flooding the excessive TCP control packet, which
might even have a catastrophic effect on the data transmission infrastructure.

In this demo, instead of using an actual heart rate
sensor, we are fabricating a procedural signal through the combination
of mathematical sine functions, and we are adding noise through the usage of a
random number generator. Each time we have a large enough batch of samples to
send, our sensor process performs a transaction for storing the new samples in
the batch. See \lstref{heart-rate-batch-transmission} for the method that performs
this batch submission transaction.

\begin{figure}[!htb]
\centering
\lstset{language=smalltalk,caption={Heart rate batch transmission transaction.}, label=\lstlabel{heart-rate-batch-transmission}}
\begin{lstlisting}[frame=single]
submitDataBatch: batchToSend
   dataStore
      withRole: role
      doTransactionWith: [ :trans |
		| sensor patient |
		sensor := role lastVersion.
		patient := sensor patient value lastVersion.
		patient heartRateSamples addAll:
			(batchToSend collect: [ :each |
				SMDHeartRateSample new
					timestamp: each timestamp;
					beatsPerMinute: each beatsPerMinute;
					yourself]).
		trans commit
	].
\end{lstlisting}
\end{figure}

\paragraph{Liveness Monitor Implementation.} The patient liveness polling monitoring
UI is implemented as a morph that performs periodical polling on the data store
with a period of $500 ms$. This polling is implemented through the automatic
stepping facilities of Morphic by overriding the \emph{step} method (See \lstref{liveness-state-polling}).
Once a new status state is received, then this UI morph is redrawn and a color
the last bpm is displayed on it. Since this transaction is only used for reading
data, it is never committed to ensure that the stored data is not changed. When
there is no explicit commit during the transaction, the transaction is aborted
by default. The explicit transaction abort in \lstref{liveness-state-polling}
is used with the objective of avoiding an error when sending
the \emph{last} message to the empty collection of samples.

\begin{figure}[!htb]
\centering
\lstset{language=smalltalk,caption={Liveness state polling transaction.}, label=\lstlabel{liveness-state-polling}}
\begin{lstlisting}[frame=single]
step
	| newStatus samples |
	newStatus := nil.
	dataStore withRole: role doTransactionWith: [ :transaction |
		samples := patient lastVersion heartRateSamples.
		samples ifEmpty: [ transaction abort ].
		newStatus := samples last.
	].

	newStatus ~= currentStatus ifTrue: [
		currentStatus := newStatus.
		self changed.
	].

\end{lstlisting}
\end{figure}

\paragraph{Medic Heart Monitoring Dashboard.} The UI for the medic that needs to
periodically check its patient  is simply a window with a single button. The action
associated to the button in this window takes care of initiating the transaction
for querying the samples. In our current version, we are simply
reading the whole data set because we have not yet implemented additional
indices that are required for optimally restricting the query into a specific
range of dates. See \lstref{heart-rate-medic-query} for the query that is used for retrieving this data.
For the final plot of the data we are simply using Roassal 3 \cite{bergel2018roassal}, a data
visualization framework.

\begin{figure}[!htb]
\centering
\lstset{language=smalltalk,caption={Heart rate data query by the medic.}, label=\lstlabel{heart-rate-medic-query}}
\begin{lstlisting}[frame=single]
fetchDataSet
	| result |
	result := #().
	self dataStore
		withRole: self medicRole
		doTransactionWith: [ :trans |
			| medic patient |
			medic := self medicRole lastVersion.
			medic patients ifEmpty: [ trans abort ].
			patient := medic patients first lastVersion.
			result := patient heartRateSamples value collect: #yourself.
	].
	^ result
\end{lstlisting}
\end{figure}

\section{Limitations}\seclabel{limitations}

\paragraph{The lack of Actors.} The main limitation in our prototype for our
data storage and processing model is the lack of a way for defining Actors. We
identify the need for at least two different kind of actors:

\begin{enumerate}
\item Autonomous actors that decide when and how to interact with the data storage model.
\item Data dependent actors, whose life cycle and operations are triggered by a condition
    met due to data changed in a transaction.
\end{enumerate}

In our heart rate scenario we are only implementing actors of the first kind in
a completely ad-hoc way. Making a proper API for streamlining the actor implementation
of this kind should not be complicated in Pharo. The interesting problem
is actually how to model actors of the second kind, and how to implement and
manage their lifecycle. It may even be desirable to automatically spawn these
actors on a separate node to the one that handled the transaction that required
spawning this actor.

\paragraph{Permissions and Authorizations.} In our current implementation, we
are using concrete roles for modeling data origins. The problem with this
mechanism is that we still need a proper way to define per-object access
control constraints and related permissions. One option for doing this is to
associate additional predicates for asking about permitted actions on objects
for a concrete role. There is also the problem on validating on whether a
concrete role does actually have the permissions for doing a certain
operations. We think that for this authorization validation problem that
association of a one-time use cryptographical token to a concrete role might
help.

\section{Threats to Validity}

\paragraph{Single Scenario Tested.} The main threat of validity for our
approach of implementing this smart data storage and processing infrastructure
is that we have only tested it in a single highly artificial and controlled test
scenario. We still have to validate our system with at least real world data,
and at least try to attempt to simulate usage scenarios that are much closer to
real world usage. Real world usage scenarios involve network partitions that
force changes on the topology of distribution, nodes that fail and are
resurrected for multiple reasons in a completely random and uncontrolled way.

\paragraph{Implementing Revocability.} Another important problem is that we still
need to address the issue of data revocation, and specially the issue of
of having to destroy data for reasons such as enforcing the privacy of
individual. We also want to be able to implement this support for data revocation
and removal in a way that does not destroy the advantages of having an append-only
storage of immutable data.

\section{Related work}

\paragraph{Git.} Git is a content addressable distributed database that is
normally used through a version control system user interface. The data storage
model used by the Git database is similar and a source of inspiration to our
data storage model. This distributed
database aspect of Git is very well hidden through this interface to be point that
most users know Git only as a distributed version control system. Internally,
Git is implemented as set of objects where a cryptographic hash of any stored
object can be used as a key for actually retrieving the object. The usage of
cryptographic hashes as keys is what makes Git content addressable database.
Directories in Git are represented as list with the hashes of the files and
subdirectories that are contained in it. A commit is  simply a tuple with
hashes to the new root directory, the parent commits, and the commit
message. Branches and tags are pointers to specific commits, which means that in
practice they are labels for the hashes of commits. 

\paragraph{Blockchain.} The blockchain is a completely distributed and
decentralized data base that was designed originally as a mechanism for
implementing the virtual currency of Bitcoin \cite{Xu17a}. The blockchain is built upon a
decentralized tree of cryptographic hashes that are replicated among all of the
nodes running on a blockchain \cite{Hari16a}. All nodes execute all transactions in a completely
deterministic way, so that all nodes can be verify the validity of the
transactions. The blockchain is also based around the concept of having an
append-only store of transactions, and this is also one of our sources of
inspiration on reusing this concept for our system. 

\paragraph{Mnesia.} The Erlang purely functional and actor based programming
language comes by default with Mnesia. Mnesia is a distributed relational
database that can store any Erlang object \cite{hebert2013learn}. Transactions
in Mnesia are represented as Erlang functions, and queries are typically encoded
as tuples that are generated via a macro transformation of an Erlang function
that encodes a boolean predicate function. We took inspiration on Mnesia for
the idea of representing transactions as blocks in our smart data prototype.

\section{Conclusions and Future Work}
In this paper we discussed the main requirements, theoretical problems, and an
architecture that seems to solve many of these problems for the construction of
a smart data storage and distribution system. We started by establishing the
requirements on having a system for storing and processing data with complete
traceability of origins, support for integrity validation, and capability of
revocating data. We also mention that any real world implementation of these
systems have to be prepared and designed to support concurrency and distributed
computing, due to the very nature of how this data is typically produced and
consumed. For these reasons we also discuss several important well-known
issues in the field of distributed computing, and distributed data bases that
at least need to be taken into account.

To mitigate the risks associated to implement this system, and to also
facilitate the support for the properties of traceability of origin, and integrity
verification, we decided to construct a model where data that is transmitted
during transactions is completely immutable, and the database only stores new data
in append-only mode. To keep the convenience of mutable state object
oriented programming, we keep ephemeral mutable copies of the deserialized objects
during the context of preparing and computing the bulk of a new transaction. But
we serialize these objects back into immutable copies that are stored in
append-only fashion at the end of a transaction.

After the description of this data model we presented a demo for our prototype
implementation for this smart data storage and processing system. This demo
presents promising results in terms of being able to express complex data
storage and processing infrastructure in a concise way by using Pharo classes. In
the future we will expand this very same demo by first modeling actors in a proper way, having much more complexity in
terms of the data processing and distribution topologies. We also think that is
crucial to add properly the concepts of actors, and actors that are triggered by
data model changes into our framework. We think that by properly modeling actors,
we might be able to remove the explicit data storage polling that is used in
our demo by a much more efficient event subscription model.



\bibliographystyle{ACM-Reference-Format}
\bibliography{references,others,rmod}


\begin{thebibliography}{20}


\ifx \showCODEN    \undefined \def \showCODEN     #1{\unskip}     \fi
\ifx \showDOI      \undefined \def \showDOI       #1{#1}\fi
\ifx \showISBNx    \undefined \def \showISBNx     #1{\unskip}     \fi
\ifx \showISBNxiii \undefined \def \showISBNxiii  #1{\unskip}     \fi
\ifx \showISSN     \undefined \def \showISSN      #1{\unskip}     \fi
\ifx \showLCCN     \undefined \def \showLCCN      #1{\unskip}     \fi
\ifx \shownote     \undefined \def \shownote      #1{#1}          \fi
\ifx \showarticletitle \undefined \def \showarticletitle #1{#1}   \fi
\ifx \showURL      \undefined \def \showURL       {\relax}        \fi
\providecommand\bibfield[2]{#2}
\providecommand\bibinfo[2]{#2}
\providecommand\natexlab[1]{#1}
\providecommand\showeprint[2][]{arXiv:#2}

\bibitem[\protect\citeauthoryear{Bergel and Mamani}{Bergel and Mamani}{2018}]%
        {bergel2018roassal}
\bibfield{author}{\bibinfo{person}{Alexandre Bergel} {and}
  \bibinfo{person}{Milton Mamani}.} \bibinfo{year}{2018}\natexlab{}.
\newblock \showarticletitle{Roassal 3}.
\newblock  (\bibinfo{year}{2018}).
\newblock


\bibitem[\protect\citeauthoryear{Black, Ducasse, Nierstrasz, Pollet, Cassou,
  and Denker}{Black et~al\mbox{.}}{2009}]%
        {Blac09a}
\bibfield{author}{\bibinfo{person}{Andrew~P. Black},
  \bibinfo{person}{St\'ephane Ducasse}, \bibinfo{person}{Oscar Nierstrasz},
  \bibinfo{person}{Damien Pollet}, \bibinfo{person}{Damien Cassou}, {and}
  \bibinfo{person}{Marcus Denker}.} \bibinfo{year}{2009}\natexlab{}.
\newblock \bibinfo{booktitle}{\emph{Pharo by Example}}.
\newblock \bibinfo{publisher}{Square Bracket Associates},
  \bibinfo{address}{Kehrsatz, Switzerland}. 333 pages.
\newblock
\showISBNx{978-3-9523341-4-0}
\urldef\tempurl%
\url{http://books.pharo.org}
\showURL{%
\tempurl}


\bibitem[\protect\citeauthoryear{Cort{\'e}s}{Cort{\'e}s}{2008}]%
        {cortes2008distributed}
\bibfield{author}{\bibinfo{person}{Jorge Cort{\'e}s}.}
  \bibinfo{year}{2008}\natexlab{}.
\newblock \showarticletitle{Distributed algorithms for reaching consensus on
  general functions}.
\newblock \bibinfo{journal}{\emph{Automatica}} \bibinfo{volume}{44},
  \bibinfo{number}{3} (\bibinfo{year}{2008}), \bibinfo{pages}{726--737}.
\newblock


\bibitem[\protect\citeauthoryear{Fischer, Lynch, and Merritt}{Fischer
  et~al\mbox{.}}{1986}]%
        {fischer1986easy}
\bibfield{author}{\bibinfo{person}{Michael~J Fischer}, \bibinfo{person}{Nancy~A
  Lynch}, {and} \bibinfo{person}{Michael Merritt}.}
  \bibinfo{year}{1986}\natexlab{}.
\newblock \showarticletitle{Easy impossibility proofs for distributed consensus
  problems}.
\newblock \bibinfo{journal}{\emph{Distributed Computing}} \bibinfo{volume}{1},
  \bibinfo{number}{1} (\bibinfo{year}{1986}), \bibinfo{pages}{26--39}.
\newblock


\bibitem[\protect\citeauthoryear{Furuhashi}{Furuhashi}{2013}]%
        {furuhashi2013messagepack}
\bibfield{author}{\bibinfo{person}{Sadayuki Furuhashi}.}
  \bibinfo{year}{2013}\natexlab{}.
\newblock \showarticletitle{MessagePack}.
\newblock \bibinfo{journal}{\emph{URL: https://msgpack. org}}
  (\bibinfo{year}{2013}).
\newblock


\bibitem[\protect\citeauthoryear{Gilbert and Lynch}{Gilbert and Lynch}{2002}]%
        {gilbert2002brewer}
\bibfield{author}{\bibinfo{person}{Seth Gilbert} {and} \bibinfo{person}{Nancy
  Lynch}.} \bibinfo{year}{2002}\natexlab{}.
\newblock \showarticletitle{Brewer's conjecture and the feasibility of
  consistent, available, partition-tolerant web services}.
\newblock \bibinfo{journal}{\emph{Acm Sigact News}} \bibinfo{volume}{33},
  \bibinfo{number}{2} (\bibinfo{year}{2002}), \bibinfo{pages}{51--59}.
\newblock


\bibitem[\protect\citeauthoryear{Han, Haihong, Le, and Du}{Han
  et~al\mbox{.}}{2011}]%
        {han2011survey}
\bibfield{author}{\bibinfo{person}{Jing Han}, \bibinfo{person}{Ee Haihong},
  \bibinfo{person}{Guan Le}, {and} \bibinfo{person}{Jian Du}.}
  \bibinfo{year}{2011}\natexlab{}.
\newblock \showarticletitle{Survey on NoSQL database}. In
  \bibinfo{booktitle}{\emph{2011 6th international conference on pervasive
  computing and applications}}. IEEE, \bibinfo{pages}{363--366}.
\newblock


\bibitem[\protect\citeauthoryear{Hari and Lakshman}{Hari and Lakshman}{2016}]%
        {Hari16a}
\bibfield{author}{\bibinfo{person}{Adiseshu Hari} {and} \bibinfo{person}{T.~V.
  Lakshman}.} \bibinfo{year}{2016}\natexlab{}.
\newblock \showarticletitle{The Internet Blockchain: A Distributed,
  Tamper-Resistant Transaction Framework for the Internet}. In
  \bibinfo{booktitle}{\emph{15th ACM Workshop on Hot Topics in Networks}}
  \emph{(\bibinfo{series}{HotNets '16})}. \bibinfo{publisher}{ACM},
  \bibinfo{address}{New York, NY, USA}, \bibinfo{pages}{204--210}.
\newblock
\showISBNx{978-1-4503-4661-0}
\urldef\tempurl%
\url{https://doi.org/10.1145/3005745.3005771}
\showDOI{\tempurl}


\bibitem[\protect\citeauthoryear{Hebert}{Hebert}{2013}]%
        {hebert2013learn}
\bibfield{author}{\bibinfo{person}{Fred Hebert}.}
  \bibinfo{year}{2013}\natexlab{}.
\newblock \bibinfo{booktitle}{\emph{Learn you some Erlang for great good!: a
  beginner's guide}}.
\newblock \bibinfo{publisher}{No Starch Press}.
\newblock


\bibitem[\protect\citeauthoryear{Maeda}{Maeda}{2012}]%
        {maeda2012performance}
\bibfield{author}{\bibinfo{person}{Kazuaki Maeda}.}
  \bibinfo{year}{2012}\natexlab{}.
\newblock \showarticletitle{Performance evaluation of object serialization
  libraries in XML, JSON and binary formats}. In \bibinfo{booktitle}{\emph{2012
  Second International Conference on Digital Information and Communication
  Technology and it's Applications (DICTAP)}}. IEEE, \bibinfo{pages}{177--182}.
\newblock


\bibitem[\protect\citeauthoryear{MongoDB}{MongoDB}{2014}]%
        {mongodb2014mongodb}
\bibfield{author}{\bibinfo{person}{Inc MongoDB}.}
  \bibinfo{year}{2014}\natexlab{}.
\newblock \showarticletitle{Mongodb}.
\newblock \bibinfo{journal}{\emph{URL https://www. mongodb. com/. Cited on}}
  (\bibinfo{year}{2014}), \bibinfo{pages}{9}.
\newblock


\bibitem[\protect\citeauthoryear{MySQL}{MySQL}{2001}]%
        {mysql2001mysql}
\bibfield{author}{\bibinfo{person}{AB MySQL}.} \bibinfo{year}{2001}\natexlab{}.
\newblock \bibinfo{title}{MySQL}.
\newblock
\newblock


\bibitem[\protect\citeauthoryear{Pritchett}{Pritchett}{2008}]%
        {pritchett2008base}
\bibfield{author}{\bibinfo{person}{Dan Pritchett}.}
  \bibinfo{year}{2008}\natexlab{}.
\newblock \showarticletitle{Base: An acid alternative}.
\newblock \bibinfo{journal}{\emph{Queue}} \bibinfo{volume}{6},
  \bibinfo{number}{3} (\bibinfo{year}{2008}), \bibinfo{pages}{48--55}.
\newblock


\bibitem[\protect\citeauthoryear{Shapiro, Pregui{\c{c}}a, Baquero, and
  Zawirski}{Shapiro et~al\mbox{.}}{2011}]%
        {shapiro2011conflict}
\bibfield{author}{\bibinfo{person}{Marc Shapiro}, \bibinfo{person}{Nuno
  Pregui{\c{c}}a}, \bibinfo{person}{Carlos Baquero}, {and}
  \bibinfo{person}{Marek Zawirski}.} \bibinfo{year}{2011}\natexlab{}.
\newblock \showarticletitle{Conflict-free replicated data types}. In
  \bibinfo{booktitle}{\emph{Symposium on Self-Stabilizing Systems}}. Springer,
  \bibinfo{pages}{386--400}.
\newblock


\bibitem[\protect\citeauthoryear{Simon}{Simon}{2000}]%
        {simon2000brewer}
\bibfield{author}{\bibinfo{person}{Salom{\'e} Simon}.}
  \bibinfo{year}{2000}\natexlab{}.
\newblock \showarticletitle{Brewer’s cap theorem}.
\newblock \bibinfo{journal}{\emph{CS341 Distributed Information Systems,
  University of Basel (HS2012)}} (\bibinfo{year}{2000}).
\newblock


\bibitem[\protect\citeauthoryear{Spec}{Spec}{[n.d.]}]%
        {specbson}
\bibfield{author}{\bibinfo{person}{BSON Spec}.}
  \bibinfo{year}{[n.d.]}\natexlab{}.
\newblock \showarticletitle{BSON (Binary JSON): Specification [Electronic
  resource]}.
\newblock \bibinfo{journal}{\emph{Mobile access: http://bsonspec. org/spec.
  html}} (\bibinfo{year}{[n.\,d.]}).
\newblock


\bibitem[\protect\citeauthoryear{Verwaest, Bruni, Lungu, and
  Nierstrasz}{Verwaest et~al\mbox{.}}{2011}]%
        {verwaest2011flexible}
\bibfield{author}{\bibinfo{person}{Toon Verwaest}, \bibinfo{person}{Camillo
  Bruni}, \bibinfo{person}{Mircea Lungu}, {and} \bibinfo{person}{Oscar
  Nierstrasz}.} \bibinfo{year}{2011}\natexlab{}.
\newblock \showarticletitle{Flexible object layouts: enabling lightweight
  language extensions by intercepting slot access}. In
  \bibinfo{booktitle}{\emph{Proceedings of the 2011 ACM international
  conference on Object oriented programming systems languages and
  applications}}. \bibinfo{pages}{959--972}.
\newblock


\bibitem[\protect\citeauthoryear{Vogels}{Vogels}{2009}]%
        {vogels2009eventually}
\bibfield{author}{\bibinfo{person}{Werner Vogels}.}
  \bibinfo{year}{2009}\natexlab{}.
\newblock \showarticletitle{Eventually consistent}.
\newblock \bibinfo{journal}{\emph{Commun. ACM}} \bibinfo{volume}{52},
  \bibinfo{number}{1} (\bibinfo{year}{2009}), \bibinfo{pages}{40--44}.
\newblock


\bibitem[\protect\citeauthoryear{Xie, Su, Kapritsos, Wang, Yaghmazadeh, Alvisi,
  and Mahajan}{Xie et~al\mbox{.}}{2014}]%
        {xie2014salt}
\bibfield{author}{\bibinfo{person}{Chao Xie}, \bibinfo{person}{Chunzhi Su},
  \bibinfo{person}{Manos Kapritsos}, \bibinfo{person}{Yang Wang},
  \bibinfo{person}{Navid Yaghmazadeh}, \bibinfo{person}{Lorenzo Alvisi}, {and}
  \bibinfo{person}{Prince Mahajan}.} \bibinfo{year}{2014}\natexlab{}.
\newblock \showarticletitle{Salt: Combining $\{$ACID$\}$ and $\{$BASE$\}$ in a
  Distributed Database}. In \bibinfo{booktitle}{\emph{11th $\{$USENIX$\}$
  Symposium on Operating Systems Design and Implementation ($\{$OSDI$\}$ 14)}}.
  \bibinfo{pages}{495--509}.
\newblock


\bibitem[\protect\citeauthoryear{Xu, Weber, Staples, Zhu, Bosch, Bass,
  Pautasso, and Rimba}{Xu et~al\mbox{.}}{2017}]%
        {Xu17a}
\bibfield{author}{\bibinfo{person}{X. Xu}, \bibinfo{person}{I. Weber},
  \bibinfo{person}{M. Staples}, \bibinfo{person}{L. Zhu}, \bibinfo{person}{J.
  Bosch}, \bibinfo{person}{L. Bass}, \bibinfo{person}{C. Pautasso}, {and}
  \bibinfo{person}{P. Rimba}.} \bibinfo{year}{2017}\natexlab{}.
\newblock \showarticletitle{A Taxonomy of Blockchain-Based Systems for
  Architecture Design}. In \bibinfo{booktitle}{\emph{IEEE International
  Conference on Software Architecture (ICSA)}}. \bibinfo{pages}{243--252}.
\newblock
\urldef\tempurl%
\url{https://doi.org/10.1109/ICSA.2017.33}
\showDOI{\tempurl}


\end{thebibliography}

\end{document}